\documentclass[letterpaper]{article} 
\usepackage{aaai24}  
\usepackage{times}  
\usepackage{helvet}  
\usepackage{courier}  
\usepackage[hyphens]{url}  
\usepackage{graphicx} 
\usepackage{natbib}  
\usepackage{caption} 
\usepackage{algorithm}
\usepackage{algorithmic}

\usepackage{newfloat}
\usepackage{listings}
\usepackage{xcolor} 

\newcommand{\bing}[1]{{\color{red}{\small\bf\sf [bing: #1]}}}

\DeclareCaptionStyle{ruled}{labelfont=normalfont,labelsep=colon,strut=off} 
\floatstyle{ruled}
\newfloat{listing}{tb}{lst}{}
\floatname{listing}{Listing}
\title{Grounding for Artificial Intelligence}
\author{
    Bing Liu
}
\affiliations{
    Department of Computer Science \\
    University of Illinois Chicago
    
    liub@uic.edu
}

\usepackage{bibentry}

\begin{document}

\maketitle

\begin{abstract}
A core function of intelligence is \textit{grounding}, which is the process of connecting the natural language and abstract knowledge to the internal representation of the real world in an intelligent being, e.g., a human. 
Human cognition is grounded in our \textit{sensorimotor experiences} in the {external world} and  subjective \textit{feelings} in our {internal world}. We use languages to communicate with each other and the languages are grounded on our shared sensorimotor experiences and feelings. Without this shard grounding, it is impossible for us to understand each other because all natural languages are highly abstract and are only able to describe a tiny portion of what has happened or is happening in the real world.  Although grounding at high or abstract levels has been studied in different fields and applications, to our knowledge, limited systematic work at fine-grained levels has been done. With the rapid progress of large language models (LLMs), it is imperative that we have a sound understanding of grounding in order to move to the next level of intelligence. It is also believed that grounding is necessary for Artificial General Intelligence (AGI). This paper makes an attempt to systematically study this problem. 
\end{abstract}

\section{Introduction}

Humans are capable of many intelligent behaviors like learning, language understanding, vision, planning, reasoning, etc. Artificial Intelligence (AI) aims to endow an AI agent, e.g., a chatbot and a physical robot, the human-level of intelligence. With the rapid progress of AI recently, especially Large Language Models (LLMs)~\cite{devlin2018bert,raffel2020exploring,brown2020language}, the community is seeing the hope of achieving Artificial General Intelligence (AGI) in a not long distance future. For AGI, grounding is a central issue \cite{bisk2020experience,assran2023self,floridi2004open,williams2009grounding}. It has been investigated in different fields, most notably philosophy~\cite{raven2015ground,rydehn2021grounding,floridi2004open} and AI and cognitive science~\cite{harnad1990symbol,vogt2007language,barsalou2008grounded,johnston2009formal,williams2009grounding}. For example,~\citet{ashwood2014linked} defines \textit{grounded knowledge} as knowledge that is contextually situated, and actively linked to other ways of knowing. \citet{williams2009grounding} defined \textit{grounding} as a process of creating representations and proposed a grounding framework. The framework comprises five essential elements that can be detailed based on the application needs: (1) system objectives, (2) architecture of grounding capability, (3) purpose and scope of the application, (4) nature of the grounding capability and (5) groundedness qualities. \citet{harnad1990symbol} proposed the \textit{symbol grounding problem}, which is concerned with the issue of how to give semantic meanings to symbols in formal systems: “How can the semantic interpretation of a formal symbol system be made intrinsic to the system, rather than just parasitic on the meanings in our heads?” However, this study is more philosophical. \citet{johnston2009formal} proposed a formal framework for the symbol grounding problem, which analyzes the space of finite formal systems and categorize them into a hierarchy of classes based on the levels of the semantic interpretability of their representations. 

All these existing works are either philosophical or highly abstract, which facilitate formal discussions of the concept.  
This paper wants to be more concrete or down-to-earth. It links grounding with  natural language words and sentences, our sensorimotor experiences, and our internal feelings because AI agents need to communicate with humans in natural languages and to understand human subjective feelings.  

We define \textit{grounding} as the process of connecting \textit{abstract knowledge} and \textit{natural language} to the internal representations of our sensorimotor experiences in the real world and our subjective feelings in our internal world. \textit{Sensorimotor experiences} can be direct or indirect. \textit{Direct sensorimotor experiences} refer to personal experiences. For example, a person sees a \textit{wolf} personally in the wild, a zoo, or a video. \textit{Indirect sensorimotor experiences} refer to situations where one is told about something by others or read the thing in a book. For example, the person has never seen a \textit{wolf} before but was told that \textit{wolfs} look like \textit{dogs} but are predators. 

In the LLM context, \textit{grounding} is defined as the process of connecting LLM answers to correct and up-to-date information \cite{potts2022could}, which usually refers to identifying creditable information sources. We call it \textit{source grounding}. Grounding in human brains is different and is significantly more complex. 
Although related, the two types of groundings are not the same because LLMs do not have snesorimotor experiences or feelings {\color{black}as they cannot explore and interact with the real-world environment by themselves and  do not have a \textit{world model} or internal \textit{emotional states} in them \cite{assran2023self}.} Most of our sensorimotor experiences cannot be easily captured or described by the language alone. Sensory grounding is thus complex~\cite{woods2007meaning} and involves more than linguistic understanding \cite{barsalou2008grounded,lake2023word}. 
We will connect the two types of groundings later. 

Although the recent progress of LLMs has prompted a great deal of discussions about grounding, the existing work on language grounding still stays at a rather high-level or is about specific tasks or applications~\cite{barsalou2008grounded,potts2022could,bisk2020experience,manning2022human,chandu2021grounding,michael2020dissect,lake2023word,bender2020climbing,masse2008meaning,patel2021mapping}. Limited fine-grained study has been done. This paper proposes a fine-grained framework for analyzing grounding in the context of natural language. 
As this is an initial attempt, there are inevitably limitations and even errors. 






\section{Proposed Grounding Framework}
\begin{figure*}[h!]
\centering
\includegraphics[width=14cm,height=12cm]{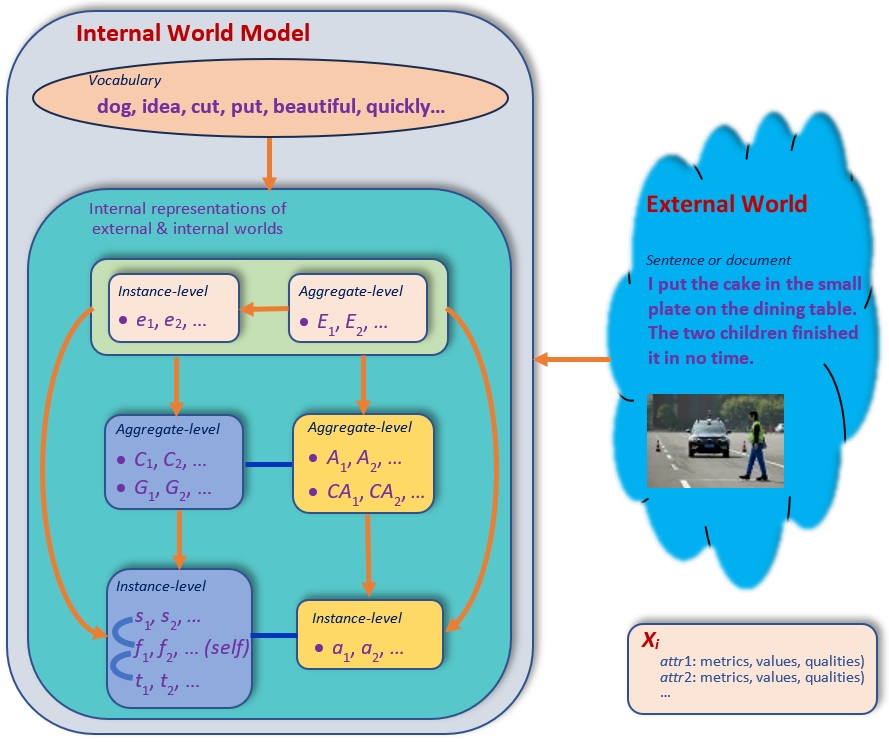}
  \caption{Grounding of external inputs in the internal representations of the world model of an intelligent being (IB). The representations can be of single-modal (e.g., visual) or multi-modal (e.g., visual and auditory). $X_i \in \{s_i, f_i, a_i, A_i, C_i, G_i, CA_i, E_i\}$ consists of a set of attributes ($attr_i$) with \textit{metrics} (or \textit{measurements}), \textit{values}, and \textit{quality-based descriptions}.}
\end{figure*}
For an intelligent being (IB) like a human, when it reads a piece of text, the content of the text is grounded in the internal representations of the model of the real world that the IB has already learned. This means that the scenario described in the text is \textit{plausible} in the real world. When the text cannot be grounded, it is due to one of the two main reasons (we assume the text has no errors):

1. There are new words in the text that the IB has not learned before and thus has no grounding of them. 

2. The scenario described in the text is not plausible based on its grounding in the IB's internal world model. For example, there are logical contradictions according to the knowledge in the IB's internal world model.\footnote{Note that this paper is not about novelty. For example, ``\textit{a man is walking a chicken in the park}'' is novel and unexpected as we usually see people walking their dogs and probably have never seen anyone walking a chicken in a park. However, walking a chicken in a park is plausible and can be grounded.} 

Whenever grounding has difficulty, the IB (e.g., a human) wants to make an effort to resolve it. For example, we can look up the new words in the dictionary to find their meanings and learn them. This involves grounding their meanings and adding them in our internal world model. We ask questions to others to try to resolve the contradictions. 

Figure 1 presents the proposed framework for the internal world model that the IB has learned and accumulated via its sensorimotor experiences in the external world and the feelings experienced in its internal world.  

\subsection{External World} 
The external world is the real world that is outside of the IB (Figure 1). It covers everything, including other IBs, that the IB can perceive through its sensory mechanisms (e.g., sight, sound or hearing, smell, taste, and touch). The perception can be \textit{direct} or \textit{indirect}.  \textit{Direct perceptions} mean personal experiences and indirect perceptions mean being told. 

\subsection{Internal World Model} 
\vspace{+1mm}
\noindent
The internal world knowledge of an IB contains the representations of the information in the external world and feelings in the internal world that the IB has experienced and learned. It is the \textbf{Internal World Model} (Figure 1).

\vspace{+2mm}
\noindent
The \textbf{instance-level} knowledge in the Internal World Model in Figure 1 contains each individual personal experience of the IB, e.g., seeing a specific object or a particular scene (with many objects) and/or experiencing a specific feeling (e.g., sadness) at a particular time. They are represented in the internal world model as follows,

$s_i$: the internal representation of an individual scene (including the particular object) that has been perceived by the IB, e.g., an image (of a particular dog), a video clip (of a particular activity), or a piece of music. 

$f_i$: the internal representation of an emotional feeling of the IB after perceiving $s_i$. \textit{Self} means the IB itself. 

$t_i$: the time when $s_i$ was perceived.  

The blue links mean that the three items are associated with one another as a triple, e.g., seeing something at a particular time causing a particular feeling. Any of the three items can be absent in a triple (e.g., ``\textit{I feel sad this morning},'' which has no specific cause). 

$a_i$: the internal representation of a particular instance of a basic action that has been performed in the external world. The blue link here means that the action may involve subjects, objects, emotional feelings, and time.  

$e_i$: the internal representation of a specific instance of an event that involves multiple specific activities and objects, e.g., a particular street demonstration. 

\vspace{+2mm}
\noindent
The \textbf{aggregate-level} knowledge in Figure 1 means the knowledge about a class or a group of objects or actions. 

$C_i$: the internal representation of a class of objects, e.g., dog or cat.  

$G_i$: the internal representation of a group of objects, e.g., a team and a committee. 

$\textit{A}_i$: the internal representation of an action class, e.g., eating and sitting

$\textit{CA}_i$: the internal representation of a composite action (e.g., buy) that consists of multiple basic actions (i.e., paying money and getting the goods). 

$E_i$: the internal representation of a class of events that involves multiple activities and objects, e.g., street demonstration (in general).  

Each item above also has a set of item-specific attributes. We use a separate box at the bottom right to specify each item. Let $X_i \in \{s_i, f_i, C_i, G_i,  a_i, A_i, CA_i\}$. It has a list of attributes ($attr_i$) (e.g., \textit{appearance},  \textit{size} and \textit{capability}), each of which has a set of associated metrics/measurements, values and qualities. The qualities are often described using adjectives or adjectives in a natural language, e.g., \textit{beautiful} for a person's appearance attribute and \textit{fast} for the speed attribute of an action.

\vspace{+2mm}
\noindent
\textbf{Vocabulary} is the list of \textbf{words} that the IB has learned and grounded in the internal representations or the \textit{world model} about the external and internal worlds. The words are mapped to their respective representations in the internal world model and are used to describe scenarios that may appear in the external world or the internal world, or to express other purposes in the communication with other IBs.

\vspace{+2mm}
\noindent
\textbf{Learning}: From instance-level experiences of a class, the learning system of the IB generalizes them to produce a model or representation of the class. For example, having seen many \textit{bananas},\footnote{Seeing here does not necessarily mean to see real bananas only. The bananas can appear in images or videos, or described by others.} the IB learns a model of the \textit{banana} class. We will not discuss leaning further in this paper as it is not the focus of the paper. 



\subsection{Grounding Rules} \label{sec.rules}
The orange arrows in Figure 1 give the direction of grounding. 
For example, when the IB reads ``\textit{I put the cake in the small plate on the dining table}'' in Figure 1, a meaning is assigned to it by grounding it in the internal world model and by showing that this activity is plausible in the real world. Specifically, ``\textit{I}'' is grounded in a person class ($C_i$) (not a specific person assuming the IB does not know who this person is), `\textit{`cake}'' is grounded in the class representation of cake ($C_j$) (not a specific ``\textit{cake}'' ($s_i$) as this was not the IB's personal experience in a specific instance), ``\textit{put}'' is grounded in the corresponding representation of the ''\textit{put}'' action class ($A_k$), and ``\textit{dining table}'' is grounded in the generic representation  of a dining table class ($C_k$). We now propose a set of grounding rules. 

\vspace{+2mm}
\noindent
\textbf{Rules of Grounding}. As defined earlier, grounding is the process of connecting a piece of text to the internal representations of objects, actions, feelings, events or scenarios described in the text. 

1. An object, feeling, action, time or event in the text that refers to a personal experience of the IB should be grounded in the instance-level representation of the experience. 

2. An object, feeling, action, time or event in the text that does not describe a personal experience of the IB should be grounded in the aggregate-level representation. This is because the IB does not have the fine-grained details, e.g., a specific visual scene. 

3. In an object, feeling, action, time or event, some aspects of it may be personal experiences of the IB and some may be not. They should be grounded at their respective levels. 

4. To save energy, unless necessary grounding goes only to the highest level that is sufficient. For example, when we read a sentence mentioning a street demonstration, the aggregate-level grounding of the event is sufficient unless the story requires lower levels of grounding of activities. 

5. To ground a full sentence, we follow the grammar to connect every word or phrase in the sentence to its corresponding representation in the internal world model (Figure 1) and simulate the scenario described in the sentence internally. Based on this grounding, if the scenario described in the sentence is plausible in the real world, the grounding is successful; otherwise it is unsuccessful.








\subsection{Grounding of LLMs}

As mentioned earlier, grounding of LLMs is the process of connecting the generated text from an LLM to the correct and up-to-date information \cite{pavlick2023symbols,potts2022could}, which we call the \textit{source grounding}. Since there is no automated method that can verify whether a piece of information is correct or up-to-date, we usually aim to identify the legitimate source of the information. Based on the proposed framework. The source of the information takes the place of the {\color{black}Internal World Model in Figure 1, but it is much simpler as it is usually just a piece of text from a creditable source. }


{Note that as human users, we are often satisfied with a reasonable answer from an LLM to our question. Some of us even believe that LLMs understand the natural language and have grounding like that in our brains. {\color{black}This may be because LLMs can generate answers that are plausible based on the grounding in our internal world model as LLMs have learned a huge amount of world knowledge that is much more than the knowledge in our individual brain.} 
}

However, the knowledge in LLMs is not grounded like the knowledge in a human brain. Clearly, LLMs do not understand what they do \cite{bender2020climbing,arcas2022large,merrill2021provable,zhang2020winowhy}. Since LLMs are statistical models, their learning methods can be applied to any language. However, for a human person, no matter how much text from a language that he/she has never learned before is given to him/her, s/he will not be to make any sense of it because human learning is mainly semantic based. To learn from a piece of text, we must first ground it in the internal world knowledge in our brain, i.e., the brain needs to make all related connections and perform all necessary reasoning or virtual simulations to ensure the plausibility of the activity or scenario described in the text before accepting it and learn from it. 


In the next section, we show/argue that all the components (including abstraction concepts) of our natural language can be grounded in our past sensorimotor experiences and subjective feelings. 


\section{Natural Language Grounding}


Natural languages evolved to facilitate high-bandwidth communication between humans. It means that in our communications, we often describe only what are important to us and leave out many fine-grained details. This does not cause any problem because we have shared cognitive frameworks and sensorimotor experiences. Those left-out details or semantic information are either unimportant or are already in our internal world models learned from our previous experiences. 

This section discusses how the components of the natural language can be grounded in our sensorimotor experiences in the external world and subjective feelings in our internal world. Below, we only discuss the word level grounding as the sentence level grounding has been discussed in point 5 of Section~\ref{sec.rules}. 

\subsection{Nouns}
Nouns are words that name people, places, things, actions, ideas, or qualities \cite{baker2003lexical}. There are several types of nouns.\footnote{The types of nouns are not mutually exclusive as the traditional categorization of nouns is based on their grammar roles and the same noun may serve multiple roles.} 

\vspace{+2mm}
\noindent
\textbf{Concrete nouns} are names of things that can be perceived with human senses (i.e., touch, hearing, sight, smell, and taste). Examples include \textit{pen}, \textit{cat}, and \textit{David}. 

\textit{Grounding}: It is easy to see that when a human sees such a noun word, it is directly grounded in the representation of the corresponding object in the internal world model (i.e., $s_i$ or $C_i$ in Figure 1). 

\vspace{+2mm}
\noindent
\textbf{Abstract Nouns} refer to things non-physical or things conceptual that we cannot perceive directly with our senses, e.g., \textit{sadness}, \textit{love}, \textit{analysis}, \textit{quality}, \textit{loyalty}, and \textit{bravery}. 

\textit{Grounding}: When a person sees nouns about feelings and emotions, they can be easily grounded in his/her internal feelings (i.e., $f_i$ or $C_i$ in Figure 1). About other abstract words like \textit{analysis} and \textit{quality}, \textit{analysis} is just the noun form of the verb \textit{analyze} and can be grounded in the name of the action \textit{analyze} (i.e., the name attribute of action $a_i$, $A_i$, or $CA_i$ under $X_i$ in Figure 1). The word \textit{quality} is just the general name for our assessment of something based on some of its specific attributes. For example, the \textit{leadership quality} of a person can be grounded in our subjective feelings about the leadership attribute of the person. \textit{loyalty} and \textit{bravery} can be grounded similarly in the quality aspects of their corresponding attributes of a person under $X_i$ in Figure 1. 

\vspace{+2mm}
\noindent
\textbf{Common nouns} name a type of person, thing, or place, e.g., \textit{student}, \textit{tree}, and \textit{dog}.

\textit{Grounding}: As its definition suggests, such a noun is simply the name of a type of objects that it can be grounded in $C_i$ in Figure 1.

\vspace{+2mm}
\noindent
\textbf{Collective nouns} name a group of people, places or things that are thought of as a single unit, e.g., \textit{team} and \textit{family}.  

\textit{Grounding}: A collective noun can be grounded in a group of objects (i.e., $G_i$ in Figure 1). 

\vspace{+2mm}
\noindent
\textbf{Proper nouns} are names for individual persons, places, or organizations, e.g., \textit{John}, \textit{Canada}, and \textit{Chicago Bull}.  

\textit{Grounding}: These nouns can be grounded in the internal representations of the specific persons, places or organizations (i.e., $s_i$ or $C_i$ in Figure 1). 

There are also other types of nouns such as compound nouns, which refer to nouns joined with two or more words together, countable nouns and uncountable nouns. Their grounding is similar to the above. 


\subsection{Verbs}

Verbs are used to describe actions \cite{baker2003lexical}. English verbs have many types, e.g., action verbs,
stative verbs,
transitive verbs,
intransitive verbs,
linking verbs,
helping verbs (also called auxiliary verbs), and 
Modal verbs \cite{leech2014meaning}. Grounding of verbs often involves subjects and objects, which are reflected by the blue links associated with actions in Figure 1  

\vspace{+2mm}
\noindent
\textbf{Action verbs} refer to actions, e.g., \textit{walk}, \textit{run}, and \textit{swim}. 

\textit{Grounding}: These verbs are directly grounded in the internal representations of their corresponding physical actions in the internal world model, i.e., $a_i$, $A_i$ or $CA_i$ in Figure 1. 

\vspace{+2mm}
\noindent
\textbf{Stative verbs} are verbs that indicate conditions or states of being. They are used to describe things like qualities, states of existence, beliefs, and emotions. Example such words include \textit{love}, \textit{want}, \textit{own} and \textit{have}.

\textit{Grounding}: These verbs can also be grounded to their respective state actions in the internal world model (i.e., $a_i$, $A_i$ or $CA_i$ in Figure 1). 


\vspace{+2mm}
\noindent
\textbf{Transitive verbs and intransitive verbs}. A transitive verb is accompanied by a direct object in a sentence. The direct object is the noun, pronoun, or noun phrase that is having something done to it by the subject of the sentence. Examples include \textit{eat}, \textit{hate}, and \textit{buy}. 

An intransitive verb is just the opposite of a transitive verb. A verb is intransitive  if it is not used with a direct object in a sentence. 

\textit{Grounding}: Both transitive and intransitive verbs can be grounded as they are either action or stative verbs but with different usages in a sentence. The blue links of $a_i$ and $A_i$ or $CA_i$ in Figure 1 reflect the associated subject and object (noun and pronoun) of either a transitive and intransitive verb. 

\vspace{+2mm}
\noindent
\textbf{Linking verbs} are a special type of stative verb whose name gives a clue as to what they do, e.g., \textit{be}, \textit{become}, \textit{seem}, and \textit{appear}.

\textit{Grounding}: They can be grounded in context. For example, ``\textit{Bert is a dog}'' assigns \textit{Bert} a category \textit{dog} which is a concrete noun. Again the blue links associated with actions in Figure 1 reflect this. 

\vspace{+2mm}
\noindent
\textbf{Helping verbs} (\textbf{auxiliary verbs})
help other verbs to change the meaning of a sentence. These include changing the tense of the verb or altering the mood of a sentence.
Example such verbs include
\textit{be}, 
\textit{have}, 
\textit{can}, and
\textit{will}.

\textit{Grounding}: They are grounded together with the main verb in a sentence. For example,``\textit{this car can move sideways}'' assigns ``\textit{moving sideways}'' to the capability attribute of the car (in $X_i$ of Figure 1).  ``\textit{moving sideways}'' is an action that can be grounded in its internal representation. 

\vspace{+2mm}
There are also other types of verbs such as regular verbs, irregular verbs
phrasal verbs, and infinitives. Their grounding are similar to the above. 

Finally, verb \textbf{tenses} can be easily grounded as well. For example, the past tense of \textit{go} in ``\textit{he went home}'' just says the action happened in the past ($t_i$ in Figure 1).

\subsection{Adjectives}

An adjective is a word that modifies a noun or a pronoun to give more information about the noun or pronoun \cite{baker2003lexical}. Common types of adjectives include 
comparative adjectives, 
superlative adjectives, 
predicate adjectives, 
compound adjectives, 
possessive adjectives, 
demonstrative adjectives, 
proper adjectives, 
participial adjectives, 
limiting adjectives, 
descriptive adjectives, 
attributive adjectives,
distributive adjectives, etc.

\textit{Grounding}: They can be grounded easily because they express different types of our subjective feelings (e.g., \textit{happy}), degrees in some metrics (e.g, \textit{tall}) or qualities (e.g., \textit{beautiful}) that we can perceive about some object attributes (e.g., \textit{appearance}) (see  qualities of attributes of $X_i$ in Figure 1). 

\subsection{Adverbs}

An adverb modifies a verb, an adjective, a clause, or another adverb. For example, the adverb \textit{fast} in the sentence ``\textit{John runs very fast}'' tells us that John runs with high speed. Common adverbs include conjunctive adverbs, 
adverbs of frequency,
adverbs of time, 
adverbs of manner,
adverbs of degree, and
adverbs of place. 

\textit{Grounding}: Similar to adjectives, adverbs can also be grounded in similar ways to subjective feelings, grades in some metrics and perceived qualities of attributes of actions (see $X_i$ in Figure 1). 

We will not discuss the other parts-of-speech as they can be grounded in their respective ways.

\section{Research Challenges and Questions}
This section discusses some research challenges and questions about grounding. By no means do we claim that this list is exhaustive or even correct. As the research progresses, the list will be improved. 

1. Perhaps the most important question is about representations. That is, what is an appropriate representation of the external world and also the internal emotional world that is flexible and powerful enough to capture the fine-grained details of the real and dynamic world environment? And, how to learn or build a world model that AI agents can use to simulate the dynamic real world environment and to exploit it for their performance tasks, e.g., reasoning and planning. Is the neural network an appropriate representation and learning framework? It seems not as neural networks are completely uninterpretable and cannot simulate the dynamic real world activities. Although the knowledge contained in human brains are also not completely interpretable, at a high level it is interpretable as our thinking is basically a process of virtually manipulating physical objects in our brain to simulate real world scenarios and object interactions in order to perform planning, reasoning, and other tasks.  

2. The internal representation must also enable incremental and continuous learning~\cite{ke2022continual}. As the world is a dynamic environment that changes constantly and is full of unknowns, the ability to continually learn and update the knowledge is critical. Although a fair amount of research has been done on continual learning in neural networks, the key challenges still remain, i.e., how to overcome forgetting of previous knowledge in learning new knowledge and how to leverage the previously learned knowledge to help learn new tasks better. Without this continual learning capability, AI agents will not be able to accumulating knowledge and working in the real world environment.    

3. Is grounding necessary for commonsense knowledge? The prevailing answer is yes. The mysterious thing is that every piece of knowledge in our brain seems to be connected to all possible related knowledge. Note that I use ``\textit{a piece}'' of knowledge here just to follow the conventional usage. I do not mean that the knowledge is discrete. Instead, the human knowledge appears to be continuous or at least is connected in a very dense graph. Whenever something is invoked, all the related pieces of information are brought to bear. Commonsense knowledge seems to sit in those connections which appear to be infinite in number. This is not surprising as the real world is infinite and ever changing. Our natural language is only able to describe a very small amount of information. The rest are dependent on our internal fine-grained representations of the complex real world.

4. As we discussed earlier, LLMs, which learn by next word prediction and generation, do not have the grounding as we described in this paper. However, they are indeed able to answer people's questions that seemly need understanding or grounding \cite{pavlick2023symbols}. As LLMs learn a great deal of knowledge from a large amount of text, some interesting questions are: how to characterize the knowledge learned in LLMs? How to ground its knowledge? And, what is the theoretical limit of LLMs? 

5. A related question is how to make an LLM aware of its knowledge. There is still no formal definition of \textit{awareness}. If we follow the prevailing definition of LLM grounding as connecting the answers from an LLM to reliable sources, we need to make LLMs aware of its knowledge so that it can verify the correctness of each piece of knowledge, find its sources, and also manipulate the knowledge to simulate hypothetical scenarios as we humans do in our thinking. In order to achieve all these, we may need to design a new way of training LLMs using multi-modal data involving audios and videos. The current LLMs are not aware of its vast amount of knowledge and cannot manipulate or update the knowledge by itself through its own interactions with humans or the external environment.


\section{Conclusion}
In this paper, we studied the problem of grounding at a fine-grained level. Although there are a great deal of discussions and studies about grounding in various fields, they are mainly philosophical or highly abstract. Detailed analysis of grounding at fine-grained levels is still lacking. As grounding is likely to be necessary for artificial general intelligence (AGI) and in order to go beyond the power of the current LLMs, it is high time to study grounding. This paper serves to encourage researchers to work on the problem. Since this is an initial study and there is little understanding on how the human brain works on grounding, I anticipate many disagreements and debates about the concepts introduced and/or discussed in this paper. If that is the case, this paper will have served its purpose.  

\section*{Acknowledgments}
This work was supported in part by four NSF grants (1910424, 1838770, 2225427, and 2229876).

\bibliography{aaai24}

\begin{thebibliography}{30}
\providecommand{\natexlab}[1]{#1}

\bibitem[{Arcas(2022)}]{arcas2022large}
Arcas, B. A.~Y. 2022.
\newblock Do large language models understand us?
\newblock \emph{Daedalus}, 151(2): 183--197.

\bibitem[{Ashwood et~al.(2014)Ashwood, Harden, Bell, and Bland}]{ashwood2014linked}
Ashwood, L.; Harden, N.; Bell, M.~M.; and Bland, W. 2014.
\newblock Linked and situated: grounded knowledge.
\newblock \emph{Rural Sociology}, 79(4): 427--452.

\bibitem[{Assran et~al.(2023)Assran, Duval, Misra, Bojanowski, Vincent, Rabbat, LeCun, and Ballas}]{assran2023self}
Assran, M.; Duval, Q.; Misra, I.; Bojanowski, P.; Vincent, P.; Rabbat, M.; LeCun, Y.; and Ballas, N. 2023.
\newblock Self-supervised learning from images with a joint-embedding predictive architecture.
\newblock In \emph{Proceedings of the IEEE/CVF Conference on Computer Vision and Pattern Recognition}, 15619--15629.

\bibitem[{Baker(2003)}]{baker2003lexical}
Baker, M.~C. 2003.
\newblock \emph{Lexical categories: Verbs, nouns and adjectives}, volume 102.
\newblock Cambridge University Press.

\bibitem[{Barsalou(2008)}]{barsalou2008grounded}
Barsalou, L.~W. 2008.
\newblock Grounded cognition.
\newblock \emph{Annu. Rev. Psychol.}, 59: 617--645.

\bibitem[{Bender and Koller(2020)}]{bender2020climbing}
Bender, E.~M.; and Koller, A. 2020.
\newblock Climbing towards NLU: On meaning, form, and understanding in the age of data.
\newblock In \emph{Proceedings of the 58th annual meeting of the association for computational linguistics}, 5185--5198.

\bibitem[{Bisk et~al.(2020)Bisk, Holtzman, Thomason, Andreas, Bengio, Chai, Lapata, Lazaridou, May, Nisnevich et~al.}]{bisk2020experience}
Bisk, Y.; Holtzman, A.; Thomason, J.; Andreas, J.; Bengio, Y.; Chai, J.; Lapata, M.; Lazaridou, A.; May, J.; Nisnevich, A.; et~al. 2020.
\newblock Experience grounds language.
\newblock \emph{arXiv preprint arXiv:2004.10151}.

\bibitem[{Brown et~al.(2020)Brown, Mann, Ryder, Subbiah, Kaplan, Dhariwal, Neelakantan, Shyam, Sastry, Askell et~al.}]{brown2020language}
Brown, T.; Mann, B.; Ryder, N.; Subbiah, M.; Kaplan, J.~D.; Dhariwal, P.; Neelakantan, A.; Shyam, P.; Sastry, G.; Askell, A.; et~al. 2020.
\newblock Language models are few-shot learners.
\newblock \emph{Advances in neural information processing systems}, 33: 1877--1901.

\bibitem[{Chandu, Bisk, and Black(2021)}]{chandu2021grounding}
Chandu, K.~R.; Bisk, Y.; and Black, A.~W. 2021.
\newblock Grounding 'Grounding' in NLP.
\newblock \emph{arXiv preprint arXiv:2106.02192}.

\bibitem[{Devlin et~al.(2018)Devlin, Chang, Lee, and Toutanova}]{devlin2018bert}
Devlin, J.; Chang, M.-W.; Lee, K.; and Toutanova, K. 2018.
\newblock Bert: Pre-training of deep bidirectional transformers for language understanding.
\newblock \emph{arXiv preprint arXiv:1810.04805}.

\bibitem[{Floridi(2004)}]{floridi2004open}
Floridi, L. 2004.
\newblock Open problems in the philosophy of information.
\newblock \emph{Metaphilosophy}, 35(4): 554--582.

\bibitem[{Harnad(1990)}]{harnad1990symbol}
Harnad, S. 1990.
\newblock The symbol grounding problem.
\newblock \emph{Physica D: Nonlinear Phenomena}, 42(1-3): 335--346.

\bibitem[{Johnston and Williams(2009)}]{johnston2009formal}
Johnston, B.; and Williams, M. 2009.
\newblock A formal framework for the symbol grounding problem.
\newblock In \emph{Conference on Artificial General Intelligence}. Atlantis Press.

\bibitem[{Ke and Liu(2022)}]{ke2022continual}
Ke, Z.; and Liu, B. 2022.
\newblock Continual learning of natural language processing tasks: A survey.
\newblock \emph{arXiv preprint arXiv:2211.12701}.

\bibitem[{Lake and Murphy(2023)}]{lake2023word}
Lake, B.~M.; and Murphy, G.~L. 2023.
\newblock Word meaning in minds and machines.
\newblock \emph{Psychological review}, 130(2): 401.

\bibitem[{Leech(2014)}]{leech2014meaning}
Leech, G.~N. 2014.
\newblock \emph{Meaning and the English verb}.
\newblock Routledge.

\bibitem[{Manning(2022)}]{manning2022human}
Manning, C.~D. 2022.
\newblock Human language understanding \& reasoning.
\newblock \emph{Daedalus}, 151(2): 127--138.

\bibitem[{Mass{\'e} et~al.(2008)Mass{\'e}, Chicoisne, Gargouri, Harnad, Picard, and Marcotte}]{masse2008meaning}
Mass{\'e}, A.~B.; Chicoisne, G.; Gargouri, Y.; Harnad, S.; Picard, O.; and Marcotte, O. 2008.
\newblock How is meaning grounded in dictionary definitions?
\newblock \emph{arXiv preprint arXiv:0806.3710}.

\bibitem[{Merrill et~al.(2021)Merrill, Goldberg, Schwartz, and Smith}]{merrill2021provable}
Merrill, W.; Goldberg, Y.; Schwartz, R.; and Smith, N.~A. 2021.
\newblock Provable limitations of acquiring meaning from ungrounded form: What will future language models understand?
\newblock \emph{Transactions of the Association for Computational Linguistics}, 9: 1047--1060.

\bibitem[{Michael(2020)}]{michael2020dissect}
Michael, J. 2020.
\newblock To dissect an octopus: Making sense of the form/meaning debate.
\newblock \emph{Blog post}.

\bibitem[{Patel and Pavlick(2021)}]{patel2021mapping}
Patel, R.; and Pavlick, E. 2021.
\newblock Mapping language models to grounded conceptual spaces.
\newblock In \emph{International Conference on Learning Representations}.

\bibitem[{Pavlick(2023)}]{pavlick2023symbols}
Pavlick, E. 2023.
\newblock Symbols and grounding in large language models.
\newblock \emph{Philosophical Transactions of the Royal Society A}, 381(2251): 20220041.

\bibitem[{Potts(2022)}]{potts2022could}
Potts, C. 2022.
\newblock Could a purely self-supervised Foundation Model achieve grounded language understanding?
\newblock \url{https://www.youtube.com/watch?v=Tp412ab3kHQ}.

\bibitem[{Raffel et~al.(2020)Raffel, Shazeer, Roberts, Lee, Narang, Matena, Zhou, Li, and Liu}]{raffel2020exploring}
Raffel, C.; Shazeer, N.; Roberts, A.; Lee, K.; Narang, S.; Matena, M.; Zhou, Y.; Li, W.; and Liu, P.~J. 2020.
\newblock Exploring the limits of transfer learning with a unified text-to-text transformer.
\newblock \emph{The Journal of Machine Learning Research}, 21(1): 5485--5551.

\bibitem[{Raven(2015)}]{raven2015ground}
Raven, M.~J. 2015.
\newblock Ground.
\newblock \emph{Philosophy Compass}, 10(5): 322--333.

\bibitem[{Ryd{\'e}hn(2021)}]{rydehn2021grounding}
Ryd{\'e}hn, H. 2021.
\newblock Grounding and ontological dependence.
\newblock \emph{Synthese}, 198(Suppl 6): 1231--1256.

\bibitem[{Vogt(2007)}]{vogt2007language}
Vogt, P. 2007.
\newblock Language evolution and robotics: issues on symbol grounding and language acquisition.
\newblock \emph{Artificial cognition systems}, 176–209.

\bibitem[{Williams et~al.(2009)Williams, McCarthy, G{\"a}rdenfors, Stanton, and Karol}]{williams2009grounding}
Williams, M.-A.; McCarthy, J.; G{\"a}rdenfors, P.; Stanton, C.; and Karol, A. 2009.
\newblock A grounding framework.
\newblock \emph{Autonomous Agents and Multi-Agent Systems}, 19: 272--296.

\bibitem[{Woods(2007)}]{woods2007meaning}
Woods, W.~A. 2007.
\newblock Meaning and links.
\newblock \emph{Ai Magazine}, 28(4): 71--71.

\bibitem[{Zhang, Zhao, and Song(2020)}]{zhang2020winowhy}
Zhang, H.; Zhao, X.; and Song, Y. 2020.
\newblock WinoWhy: A deep diagnosis of essential commonsense knowledge for answering Winograd schema challenge.
\newblock \emph{arXiv preprint arXiv:2005.05763}.

\end{thebibliography}

\end{document}